# RESOURCE-OPTIMAL PLANNING FOR AN AUTONOMOUS PLANETARY VEHICLE

Giuseppe Della Penna[1], Benedetto Intrigila[2] , Daniele Magazzeni[3] and Fabio Mercorio[1]

[1]Department of Computer Science, University of L'Aquila, L'Aquila – Italy
giuseppe.dellapenna@univaq.it, fabio.mercorio@univaq.it
[2]Department of Mathematics, University of Rome "Tor Vergata", Rome – Italy
intrigil@mat.uniroma2.it
[3]Department of Science, University of Chieti, Chieti – Italy
magazzeni@sci.unich.it

### *ABSTRACT*

*Autonomous planetary vehicles, also known as rovers, are small autonomous vehicles equipped with a variety of sensors used to perform exploration and experiments on a planet's surface. Rovers work in a partially unknown environment, with narrow energy/time/movement constraints and, typically, small computational resources that limit the complexity of on-line planning and scheduling, thus they represent a great challenge in the field of autonomous vehicles. Indeed, formal models for such vehicles usually involve hybrid systems with nonlinear dynamics, which are difficult to handle by most of the current planning algorithms and tools. Therefore, when offline planning of the vehicle activities is required, for example for rovers that operate without a continuous Earth supervision, such planning is often performed on simplified models that are not completely realistic. In this paper we show how the UPMurphi model checking based planning tool can be used to generate resource-optimal plans to control the engine of an autonomous planetary vehicle, working directly on its hybrid model and taking into account several safety constraints, thus achieving very accurate results.*

## 1. INTRODUCTION

Autonomous planetary vehicles, or rovers, are a great challenge in the field of autonomous vehicles, since they have often to take actions on a hazardous ground with narrow time and energy consumption constraints. Rovers operating on distant planets may receive commands from Earth operators only once per day, and during the remaining time they have to perform a specific mission, which may include moving to a specific place, position some instruments, take measures, etc [1], [2].

Usually, rover activities are organised on the ground into a detailed plan that, once generated and uploaded to the vehicle, drives it for the rest of the mission. Typically, in this situation, if the rover encounters an unexpected (i.e., not contained in the mission plan) situation, it stops and waits for further Earth instructions, possibly wasting a lot of time. Therefore, planning for this kind of autonomous vehicles should be very precise and take into consideration many factors [3].

As noted, many rover activities begin with a movement that places it in a specified location. Thus, independently from the nature of the rover's mission, reaching the activity location is the first goal to achieve, and it must satisfy two main constraints: energy and time consumption. Indeed, at the end of the journey, rover batteries must still contain enough charge to perform the activity, or at least to begin it, assuming that the vehicle is equipped with solar panels and in sunlight, so it can recharge its battery when needed. Moreover, the rover has to complete its task as soon as possible, since environmental conditions may quickly change, and in general a shorter task duration means that the rover will be able to perform more activities during the mission time.

Unfortunately, also in this simplified setting, the rover still represents a hybrid system (i.e., described by both discrete and continuous variables), and the equations describing values such as energy consumption are usually nonlinear. Many efforts have been made to improve planning algorithms so as to cope with hybrid systems [4], however nonlinearity continues to be a relevant problem for many well-known planners such as TM-LPSAT [5], UPPAAL-TIGA [6] or MIPS [7]. Therefore, while a precise planning of the rover activities is important, automatically generating an optimal plan w.r.t. time and energy for the rover movements may be a hard task.

### 1.1. Contribution

In this paper, which is an extended version of [8], we show how an explicit model checking-based planner, namely UPMurphi [9], developed by the same authors, can be used to generate optimal plans to control a rover's engine, in order to move it for a specific distance in the least possible time, while satisfying a set of technical constraints and trying to save energy.

Indeed, thanks to the model checking algorithms, UPMurphi is able to cope with systems showing high complexity and a behaviour that is difficult to predict, as hybrid nonlinear systems usually are.

In the presented case study, the rover dynamics and behaviour, including some common technical constraints, have been modelled through general equations, that may apply to a wide range of vehicles. Plans have been optimised to minimise energy and time requirements, and the given minimal battery charge is always preserved. Therefore, the results are quite realistic.

To the best of our knowledge, this is the first successful attempt to automatically generate time and resource-optimal plans for a rover model of such complexity.

The paper is organised as follows. Section 2 shows some related work about planning algorithms and tools for hybrid and nonlinear systems. Section 3 describes the model checking approach to the planning problem and introduces the UPMurphi planner. Section 4 defines our case study and comments the planning results. Finally, Section 5 contains some closing remarks.

## 2. RELATED WORK

The planning problem presented in this paper can be generally defined as *planning with resource consumption* [10], [11], [12].

Different techniques have been developed by the JPL [13] and applied to such kind of activity planning for NASA rovers. In particular, *the mixed initiative planning* [14], supported by the MAPGEN (Mixed-initiative Activity Plan GENerator) tool [15], [16] has been used for the Deep Space 1 mission. With this approach, humans and machines collaborate in the development and the management of plans: in particular, the user provides MAPGEN with a qualitative evaluation of the generated plan.

The ASPEN [17], [18] framework has been used to perform activity planning for the Rocky 7 rover. In this framework, a first automatically generated plan is iteratively refined using different heuristics to finally fulfil the resource constraints.

Finally, other approaches perform *planning with uncertainty*, mainly using probabilistic methods [12], which however cannot handle problems with high complexity.

It is worth noting that none of the above approaches is able to find *optimal* plans, i.e., plans that minimise the resource consumption. Other well known automatic planners indeed address this issue. In particular, the TM-LPSAT tool [5] works only on linear domains. This is also the case of the UPPAAL/TIGA tool [6], [19] that, being built on top of UPPAAL, allows one to use real variables only as clocks (i.e., real variables can be modelled only if their first derivative is 1), thus excluding systems with nonlinear dynamics. Finally, the MIPS planner [7], being based on symbolic model checking, does not perform well on hybrid nonlinear systems.

Therefore, automatic optimal planning with resource consumption on hybrid, nonlinear systems such as rovers is still an open issue.

## 3. Planning through Explicit Model Checking

Model checking algorithms are typically divided in two categories: symbolic algorithms (e.g., [20]) and explicit algorithms (e.g., [21]). Symbolic algorithms have been successfully applied to classical planning [7,22], however they do not work well on hybrid systems with nonlinear dynamics, due to the complexity of the state transition function [23]. Therefore, explicit model checking performs better with the kind of systems we intend to approach. Also these algorithms are subject to the well-known state explosion problem: however, the ability to build the system transition graph on demand and generate only the reachable states of the system (through the reachability analysis), together with many space saving techniques (see, e.g., [24]), help to mitigate it.

Generally speaking, given a set $E$ of error states and a set $I$ of *initial states* for a system, an explicit model checker incrementally generates all the valid system states *(reachable states)*, starting from the ones in $I$ and repeatedly applying the *transition function* that describes the system dynamics. If a state $e \in E$ is encountered, the model checker outputs the sequence of states *(error trace)* that leads to $e$.

If we look at error states as *goal states*, we can use a model checker as a planner. This very simple fact allows one to use the model checking technology to automatically synthesise optimal plans for complex systems.

### 3.1. Planning of Finite State Systems

A hybrid system [25] is a system whose state description involves continuous as well as discrete variables. In order to apply model checking algorithms and exploit the reachability analysis, the system should have a finite number of states. To this aim, we approximate the system by discretising the continuous components of the state (which we assume to be bounded) and their dynamic behaviour. For lack of space, we cannot describe here the approximation process, however the reader can see how our approach works by looking at the case study in Section 4. In the following we first give a formal definition of the approximated model, *the finite state system*, and then describe how the planning problem can be solved for such kind of system.

**Definition 1 (Finite State System)** A Finite State System (FSS) $\mathcal{S}$ is a 4-tuple $(S, s_0, A, F)$, where: $S$ is a finite set of states, $s_0 \in S$ is the initial state, $A$ is a finite set of actions and $F: S \times A \times S \to \{0,1\}$ is the transition function, i.e. $F(s, a, s') = 1$ iff the system from state $s$ can reach state $s'$ via action $a$.

By abuse of language, we denote with $F(s, a)$ the successor state of $s$ through action $a$, i.e. the state $s'$ so that $F(s, a, s') = 1$.

In order to define the planning problem for such a system, we assume that a set of *goal states* $G \subseteq S$ has been specified. Moreover, to have a finite state system, we fix a *finite temporal horizon* $T$ and we require a plan to reach the goal in at most $T$ actions. Note that, in most practical applications, we always have a maximum time allowed to complete the execution of a plan, thus this restriction, although theoretically quite relevant, has a limited practical impact.

Now we are in position to state the planning problem for finite state systems.

**Definition 2 (Planning Problem on FSS)** Let $\mathcal{S} = (S, s_0, A, F)$ be an FSS. Then, a planning problem is a quadruple $P = (\mathcal{S}, G, C, T)$ where $G \subseteq S$ is the set of the goal states, $C: S \times A \to \mathbb{R}^+$ is the cost function and $T$ is the finite temporal horizon.

Intuitively, a solution to a planning problem is the minimal cost path in the system transition graph, starting from the initial state and ending in a goal state. More formally, we have the following

**Definition 3 (Trajectory)** A trajectory in the FSS $\mathcal{S} = (S, s_0, A, F)$ is a finite sequence $\pi = s_0 a_0 s_1 a_1 s_2 a_2 \ldots a_{n-1} s_n$ where $s_i \in S$ is a state, $a_i \in A$ is an action and $\forall i \in [0, n-1]$ $F(s_i, a_i, s_{i+1}) = 1$. If $\pi$ is a trajectory, we write $\pi_s(k)$ (resp. $\pi_a(k)$) to denote the state $s_k$ (resp. the action $a_k$). Finally, we denote with $|\pi|$ the length of $\pi$, given by the number of actions in the trajectory, and $C(\pi) = \sum_{i=0}^{|\pi|-1} C(s_i, a_i)$. the cost of $\pi$.

**Definition 4 (Reachable States)** Let $\mathcal{S} = (S, s_0, A, F)$ be an FSS. Then, we say that $s \in S$ is reachable from $s_0$ iff there exists a trajectory $\pi$ in $\mathcal{S}$ such that $\pi_s(0) = s_0$ and $\pi_s(k) = s$ for some $k \geq 0$. We denote with $\mathrm{Reach}(s)$ the set of states reachable from $s_0$.

**Definition 5 (Admissible Solution)** Let $\mathcal{S} = (S, s_0, A, F)$ be an FSS and let $P = (\mathcal{S}, G, C, T)$ be a planning problem. Then an admissible solution for $P$ is a trajectory $\pi^*$ in $\mathcal{S}$ s.t.: $|\pi| = k$, $k \leq T$, $\pi_s^*(0) = s_0$ and $\pi_s^*(k) \in G$.

**Definition 6 (Optimal Solution)** An optimal solution is an admissible solution $\pi^*$ s.t. for all other admissible solutions $\pi'$, $C(\pi^*) \leq C(\pi')$.

In the next section, we present a tool which takes as input a planning problem and outputs an optimal solution for it.

### 3.2. The UPMurphi Tool

The UPMurphi tool [9] is an *optimal universal planner* built on top of the CMurphi [26] model checker. Obviously, UPMurphi can be used as an *optimal planner*, too.

UPMurphi can be fed with a description of the system to be verified, defined through the *UPMurphi description language* (as a finite state automaton) as well as through PDDL+ [27], which is a standard language for planning problems. This allows one to model the planning domain using the formalism that best suits the original system specifications and simplifies its description. Similarly, plans computed by UPMurphi can be output as binary or textually-encoded state-action tables or as PDDL+ plans.

The case study of this paper, being developed from scratch, has been directly modelled through the CMurphi description language, a high-level programming language for finite state asynchronous concurrent systems which allows greater flexibility and optimisation of the model.

In particular, the system state is defined in the UPMurphi through a set of *state variables*, which are suitably declared at the beginning of the model description. To this aim, the user can exploit the built-in data types provided by the tool or declare new user-defined types using data definition primitives such as arrays, structures and ranges.

The behavioural part of an UPMurphi model is a collection of *transition rules*, i.e., guarded commands which consist of a *condition* and an *action*. It is also possible to specify a *duration* and a *weight* (e.g., a cost) in each rule: these properties can be later used by the planner to measure the total duration of a plan and optimise it w.r.t. its total weight. Indeed, plans generated by UPMurphi are actually represented by a timed sequence of rules chosen by the planner to reach a goal state starting from the initial state. It is also possible to write support functions and procedures and call them in the condition, action, duration or weight expressions to further modularise the specification.

Error conditions can be defined by means of the *invariant* construct, which allows one to describe the constraints that must be fulfilled by every system state and, as a consequence, the states that represent system errors.

The system initial state is declared through the *startstate* construct, which requires the user to suitably initialise all the model state variables. Finally, to support planning problems, the CMurphi input language has been extended in UPMurphi to include the *goal* construct, used to define the properties that a planning goal state must satisfy.

Moreover, UPMurphi provides two important features to ease the hybrid systems modelling activity: the type *real(m,n)* of real numbers (with *m* digits for the mantissa and *n* digits for the exponent), and the use of externally defined C/C++ functions in the modelling language. In this way, for example, one can use the C/C++ language constructs and library functions to model complex dynamics involving calculations on real values.

Finally, UPMurphi can exploit several techniques that help to mitigate the well-known state explosion problem due to the memory requirements of explicit state space exploration algorithms. However, it is worth noting that memory-related problems are common to all the current planning algorithms, when the system dynamics involves real values and complex mathematical operations.

In particular, the tool supports *bit compression* [28] and *hash compaction* [29], [30] to reduce the memory size of the system state representation, and a symmetry reduction algorithm to decrease the state space size by detecting equivalent states.

Shortly, bit compression saves memory by using every bit of the *state descriptor*, the memory structure maintaining the state variables, instead of aligning them on byte boundaries, whereas hash compaction stores compressed values (also called *state signatures*) instead of full state descriptors. Together, these two techniques can dramatically reduce the memory needed to explore huge state spaces.

## 4. CASE STUDY: THE AUTONOMOUS PLANETARY VEHICLE

This section presents a motivating case study where planning is applied to automatically control the engine of an autonomous vehicle during a planetary exploration mission.

As described in the introduction, the rover model as well as the environmental conditions have been defined to be as general as possible, to achieve realistic results, and the generated plan will be optimised for the shortest time and the lowest energy consumption, as a real mission would require.

### 4.1. Rover Specifications

The rover can be naturally modelled as a hybrid system, with several nonlinear characteristics. Thus, we have a dynamics very hard to compute, which makes planning quite difficult. The rover model used in our case study is based on the Mars exploration rover described in [31].

In general, an exploration rover moves on the planet surface to observe different phenomena and/or try some experiments. The rover can recharge its batteries through a solar panel, but recharge cannot take place continuously, and the energy from the panels is not enough to directly power the rover. Therefore, it must minimise the energy consumption in order to have always enough battery charge for the next activity.

Moreover, the rover has limited communication and computation resources, so it must be programmed with a detailed plan of activity and then left operating, without any chance to recover from an error or recompute its mission. If something wrong or unexpected happens, the best that the rover can do is to stop, reset and wait for the next Earth connection to get new instructions.

The plan we want to generate does not address the actual route of the rover, but controls the vehicle engine and instruments during the route itself. Routing is a different problem, so just we assume that a (possibly straight) route of length $d_{final}$ has been separately planned and will be used to control the steering of the rover wheels.

When moving, the rover is subject to friction and drift due to the - often unpredictable - ground characteristics. Thus, every $d_{max}$ meters, it has to stop for $t_c$ seconds to look at its actual position and conditions, before starting again to move. These frequent stops may also be useful to ensure a proper cooling of the rover wheels and instruments, if moving in a hot environment. For sake of generality, in the following we shall call these stops "cooling tasks". However, we assume that the route duration be less or equal to $t_{max}$ seconds, since the overall rover mission should not exceed a reasonable limit. The rover has a base energy consumption $g_s$ Joule/second, used to power its CPU.

The energy (expressed in Joule/second) required to move the rover with speed $v$ and acceleration $\dot{v}$ can be evaluated by applying the general function $f$ of Equation 1, where $m$ is the vehicle mass and $fa$ is its frontal area (see [31] for details).

$$f(v,\dot{v}) = (\frac{1}{2} \cdot \rho \cdot v^2 \cdot Cd \cdot fa + m \cdot g \cdot (Crr + \frac{\dot{v}}{g})) \cdot v \quad (1)$$

In the equation, constants , $g$ indicate the planet air density and its gravitational constant, respectively, whereas $Cd$ and $Crr$ are the drag and rolling coefficients of the rover. Finally, the cooling tasks require a constant energy of $g_c$ Joule/second.

The rover dynamics (i.e., the covered distance $d$, the speed $v$ and the acceleration $\dot{v}$ ) is given by Equation 2.

$$\begin{aligned} \frac{\vartheta v}{\vartheta t} &= a(t) - \mu \cdot g \\ \frac{\vartheta d}{\vartheta t} &= v(t) \end{aligned} \quad (2)$$

where $a(t)$ is the acceleration given by the rover motor at time $t$ and $\mu$ is the kinetic friction coefficient for the rover wheels.

We assume that, in each communication session, the Earth control sends to the rover a plan to drive it to the next place, and the commands needed to start the corresponding activity. Such plan consists of a sequence of actions, to be performed at 1 second intervals, chosen from the set A = {*accelerate, decelerate, continue* (moving at constant speed), perform a *cooling task*}.

The plan must obey the following constraints:

- the rover must not exceed the speed of $v_{max}$ cm/s;
- the rover must stop every $d$ meters to perform a cooling task;
- the rover must stop after $d_{final}$ meters (to start the activity) with a residual battery charge not lower than $c_{min}$ coulomb;
- the rover route must not require more than $t_{max}$ seconds.

In particular, we must ensure that, after moving to the given location, the rover has still enough battery charge available for its activity.

## 4.2. Rover Modelling

The dynamics and constraints given above can be compactly and precisely illustrated through the hybrid automaton shown in Figure 1.

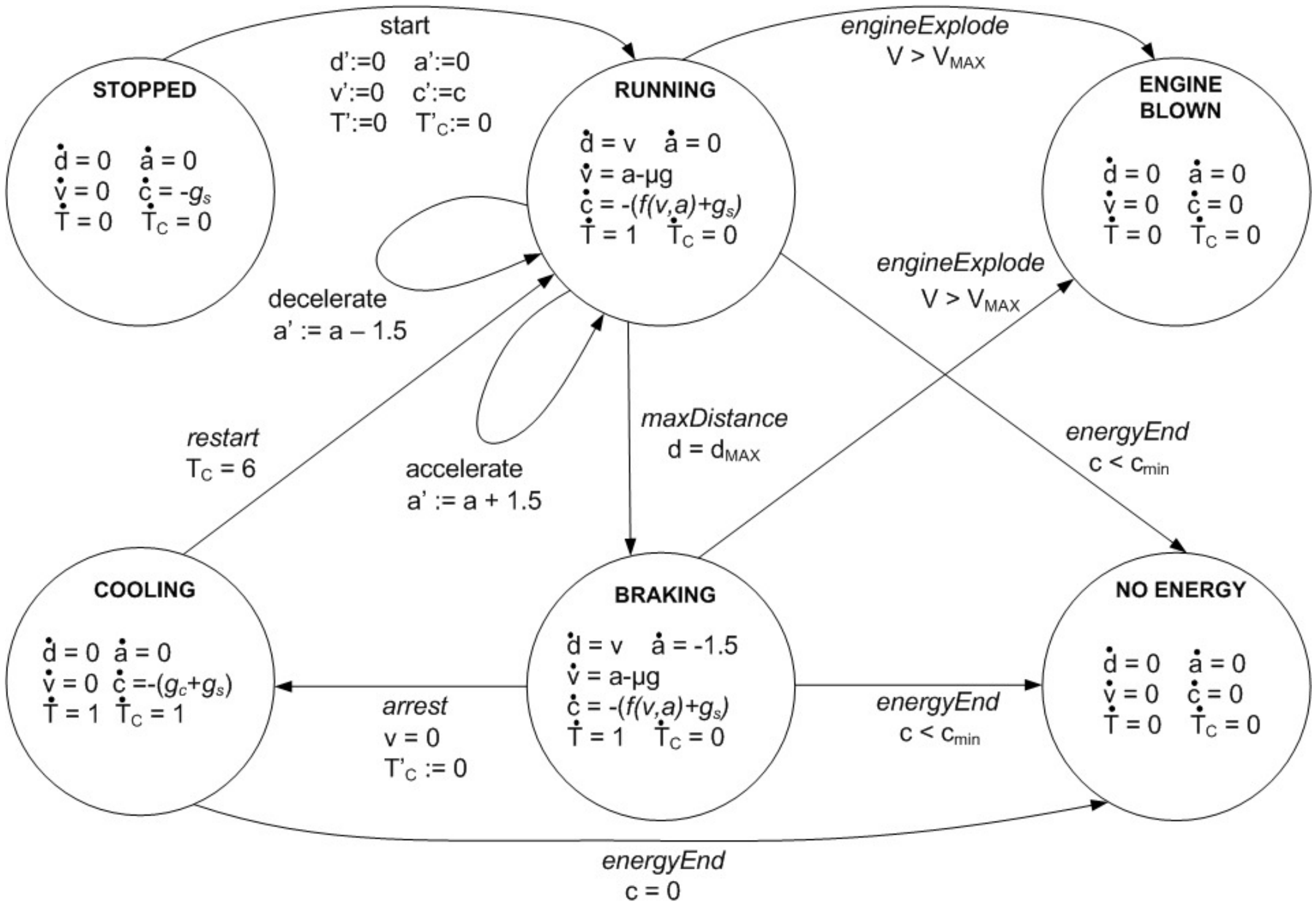

Figure 1. Hybrid automaton for the Autonomous Planetary Vehicle case study.

The state of the automaton is $s = (x, q) \in \mathcal{S}$, where $q \in$ {stopped, running, braking, cooling, engine blown, no energy} and $x = (d, a, v, T, T_c)$.

The rover is initially in a *stopped* state, where the only energy consumption is given by $g_s$. When started, the rover enters the *running* state and moves as described by Equation 2 while its energy consumption is increased by the value given by Equation 1. The vehicle can accelerate and decelerate with steps of $1.5\ cm/s^2$. After $d$ meters, the vehicle starts *braking* and, once stopped, it begins the *cooling* phase, with the corresponding energy consumption. After 6 seconds of cooling ($T_c$ in the automaton), the vehicle restarts and continues in the *running* state.

The automaton also shows two possible failure conditions: if the rover moves faster than the max allowed speed $v_{max}$ , its engine blows up (*engine blown* state): in this case, the entire mission could fail. On the other hand, if the consumed energy exceeds the limit $c_{min}$ , the rover stops *(no energy state)*, using the residual energy to wait for Earth instructions.

We fixed the model constants to the values given in Table 1, most of which are obtained from rover specifications like [2] and [1]. Note that we assume that the rover operates on the Mars surface.

Table 1. Constant values for the rover model

| | | |
|---|---|---|
| $\rho$ | Air density | 0.1 $Kg/m^3$ |
| $g$ | Gravitational acceleration | 3.8 $m/s^2$ |
| $m$ | Vehicle mass | 71.73 $Kg$ |
| $\mu$ | Kinetic friction coefficient | 0.8 |
| $c_{max}$ | Initial battery charge | 18,000$C$ |
| $c_{min}$ | Min final battery charge | 17,000$C$ |
| $v_{max}$ | Max speed | 1.0 $cm/s$ |
| $a_{max}$ | Max acceleration | 5 $cm/s^2$ |
| $g_s$ | CPU energy requirements | 25 $J/s$ |
| $t_c$ | Cooling duration | 6 s |
| $d_{max}$ | Distance between coolings | 1.30 $m$ |
| $g_c$ | Cooling energy requirements | 10 $J/s$ |
| $d_{final}$ | Final distance | 2 $m$ |
| $t_{max}$ | Max plan duration | 60$s$ |

Finally, according to Definition 3, we evaluate the cost of the generated plan through the function $C(s_i, a_i)$ defined as follows:

$$C(s_i, a_i) = \begin{cases} \dfrac{{g_s}^2}{t_{max} - i} + C_a(a_i) & if \quad q_i = stopped \\ \dfrac{(g_s + g_c)^2}{t_{max} - i} + C_a(a_i) & if \quad q_i = cooling \\ 0 & if \quad q_i \in \left\{ \begin{matrix} no\ \ energy, \\ engine\ \ blown \end{matrix} \right\} \\ \dfrac{(g_s + f(v_i, \dot{v}_i))^2}{t_{max} - i} + C_a(a_i) & otherwise \end{cases}$$

where $s \in S,\ a \in A$. Here, $C_a = 0$ since all the actions are instantaneous and do not require energy. This definition of $C$ allows one to perform optimisation on both energy and time, as required, still giving more importance to the energy component. Indeed, usually the mission could be accomplished even if it requires some seconds more than the planned limits, whereas running out of battery charge could lead to dangerous failures.

### 4.3. UPMurphi Model

The resulting model has been translated to a FSS, encoded in the UPMurphi description language, with the same state variables and transition function of the hybrid automaton in Figure 1.

In this phase, the continuous state variables have been suitably discretised. It is worth noting that this discretisation is indeed *realistic*, since also the real rover instruments, being built on a limited hardware (e.g., sensors, actuators), would implicitly work on approximations of the real continuous values. However, we must be aware of approximations that may lead to unexpected violations of safety constraints.

Indeed, in the UPMurphi rover model, we applied an approximation of 0.1 to all the variables, and introduced the $v_{safemax} \leq v_{max}$ constant as the actual maximum speed. This gives us a chance to set a further safety threshold on the speed, to prevent an engine blow (see Section 4.2) due to approximation errors. On the other hand, the journey time will be measured in seconds, since it is a reasonable update interval for the rover engine status. It is worth noting that, with the given discretisation, the total number of different states of the FSS is $2.2 \cdot 10^{13}$. Figure 2 shows the resulting UPMurphi code, where for sake of simplicity we omit the declaration of constants and state variables.

**Model rules**

```
startstate " stopped "
BEGIN
  a := 0.0;  d := 0.0;
  v := 0.0;  c := c_max;
  T_c := 0.0;
  cooling := false;
  braking := false;
  running := false;
END;

rule " start "
duration: 0;
weight: 0;
(!running & !cooling & !braking ) ==>
BEGIN
running :=true;
END;

rule " accelerate "
duration: 0;
weight: 0;
(running & !cooling & !braking ) ==>
BEGIN
a := a + 1.5;
END;

rule " decelerate "
duration: 0;
weight: 0;
(running & !cooling & !braking) ==>
BEGIN
a := a - 1.5;
END;

rule " running "
duration: 1;
weight: cost_moving();
(running & !cooling & !braking ) ==>
BEGIN
running_status_update();
END;

rule " braking "
duration: 1;
weight: cost_moving();
(!running & !cooling & braking ) ==>
BEGIN
braking_status_update();
END;

rule " cooling "
duration: 1;
weight: cost_cooling();
(!running & cooling & !braking) ==>
BEGIN
cooling_status_update();
END;

invariant " engineExplode "
(!(running & v > v_safemax))

invariant " energyEnd " (!(c < c_min))

goal " success " (v = 0 & d = d_final)
```

**Support procedures**

```
procedure running_status_update();
BEGIN
  d := update_d(d,v,a);
  v := update_v (v,a);
  c := update_c(rho,v,m,g,a,h,f);

  -- maxDistance
  IF ((d = d_max) & (T_C = 0)) THEN
    braking := true;
    running := false;
  ENDIF;
END;

procedure braking_status_update();
BEGIN
  a := a - 1.5;
  d := update_d (d,v,a);
  v := update_v (v,a);
  c := update_c (rho,v,m,g,a,h,f);

  -- arrest
  IF (v=0 & a=0) THEN
   braking:= false;
   cooling:= true;
    T_c := 0;
  ENDIF;
END;

procedure cooling_status_update();
BEGIN
  T_c := T_c +1;

  -- cooling
  IF (T_c <= 6) THEN
    c := update_c_cooling(c,g,v,m);
  ELSE
    -- restart
    cooling := false;
    running := true;
  ENDIF;
END;
```

Figure 2. UPMurphi code for the Autonomous Planetary Vehicle case study.

The start state of the model, *stopped,* describes the corresponding initial state of the hybrid automaton, i.e., fixes the initial conditions of the rover. Then, the *start* rule initiates the rover movement by setting the *running* variable to true.

The other five model rules, namely *accelerate*, *decelerate*, *running, braking* and *cooling*, model the main transitions and states of the automaton. In particular, *accelerate* and *decelerate* update the acceleration variable as described by the corresponding automaton transitions. These rules have a null duration and weight, according to the hybrid automaton semantics, since they represent instantaneous updates.

On the other hand, the *running, braking* and *cooling* rules have *duration* 1, since they model the changes in the rover state (i.e., speed, distance and battery charge) during a time step of one second. Such updates are actually performed by the *running_status_update, braking_status_update* and *cooling_status_update* procedures, respectively, which concentrate the update logic found in the entire automaton, i.e., the updates specified on the *maxDistance*, *arrest* and *restart* transitions and the ones contained in the *running* and *cooling* states. The status update procedures, in turn, compute some values through external C functions (e.g., *update_c_cooling*) that are used to evaluate the complex expressions described in Section 4.1. Moreover, the external functions *cost_moving* and *cost_cooling* are used to dynamically calculate the *weight* of each rule, as defined by the cost function shown in Section 4.2.

The invariants *engineExplode* and *energyEnd* model the homonymous transitions that lead, in the automaton, to error states (*engineBlown* and *noenergy*, respectively). These states are not modelled here, since the planner automatically detects as errors all the states that violate an invariant.

Finally, the *goal* construct is used to declare the success condition of the model, i.e., when the rover completes successfully its journey.

### 4.4. Planning

To build the optimal plan, the model was given in input to UPMurphi. Initially, we set $v_{safemax} = v_{max}$, to see if we can devise a safe plan without imposing more restrictive constraints on the rover speed. The exploration of the model dynamics lead to 939,477 reachable states, which is considerably smaller than the theoretical system state space of $2.2 \cdot 10^{13}$ states. Thanks to such state space pruning, due to the reachability analysis performed by the model checking algorithms of UPMurphi, finding an optimal (w.r.t. the cost function) plan required a relatively small amount of resources, with a peak memory allocation of 500 MB, and 2,257 seconds of processing.

The resulting plan is described in Table 2. The table reports, for each second (which is the plan sampling time, as discussed earlier) the model rule (with respect to the code in Figure 2) chosen by UPMurphi. Thus, the rover starts its journey when *Start* is selected, moves when *Running* is selected, brakes when *Braking* is selected, increases or decreases its speed when *Accelerate* or *Decelerate* are selected, respectively, and performs a *Cooling* when the homonymous rule is chosen. Note that we may have more than one rule executed in a single time step, since some of them (namely, *Start*, *Accelerate* and *Decelerate)* have duration zero.

Table 3 summarises the plan statistics. It is worth noting that the plan optimisation allowed us to save 922.7 C with respect to the required minimal battery charge, and 17 seconds with respect to the maximum allowed plan duration.

Table 2. Optimal plan

| **T** ***(sec)*** | **Rule** | **T** ***(sec)*** | **Rule** | **T** ***(sec)*** | **Rule** |
|---|---|---|---|---|---|
| 0 | Start Accelerate Running | 15 | Running | 30 | Cooling |
| 1 | Accelerate Running | 16 | Running | 31 | Cooling |
| 2 | Running | 17 | Running | 32 | Cooling |
| 3 | Decelerate Running | 18 | Running | 33 | Cooling |
| 4 | Decelerate Running | 19 | Running | 34 | Cooling |
| 5 | Decelerate Running | 20 | Running | 35 | Accelerate Running |
| 6 | Running | 21 | Running | 36 | Accelerate Running |
| 7 | Running | 22 | Accelerate Running | 37 | Running |
| 8 | Accelerate Running | 23 | Running | 38 | Decelerate Running |
| 9 | Running | 24 | Running | 39 | Decelerate Running |
| 10 | Running | 25 | Braking | 40 | Decelerate Running |
| 11 | Running | 26 | Braking | 41 | Decelerate Running |
| 12 | Running | 27 | Braking | 42 | Decelerate Running |
| 13 | Running | 28 | Braking | | |
| 14 | Running | 29 | Cooling | | |

Table 3. Optimal plan statistics

| | |
|---|---|
| Course length | $43s$ |
| Energy consumption | $77.3C$ |
| Residual battery charge | $17{,}922.7C$ |

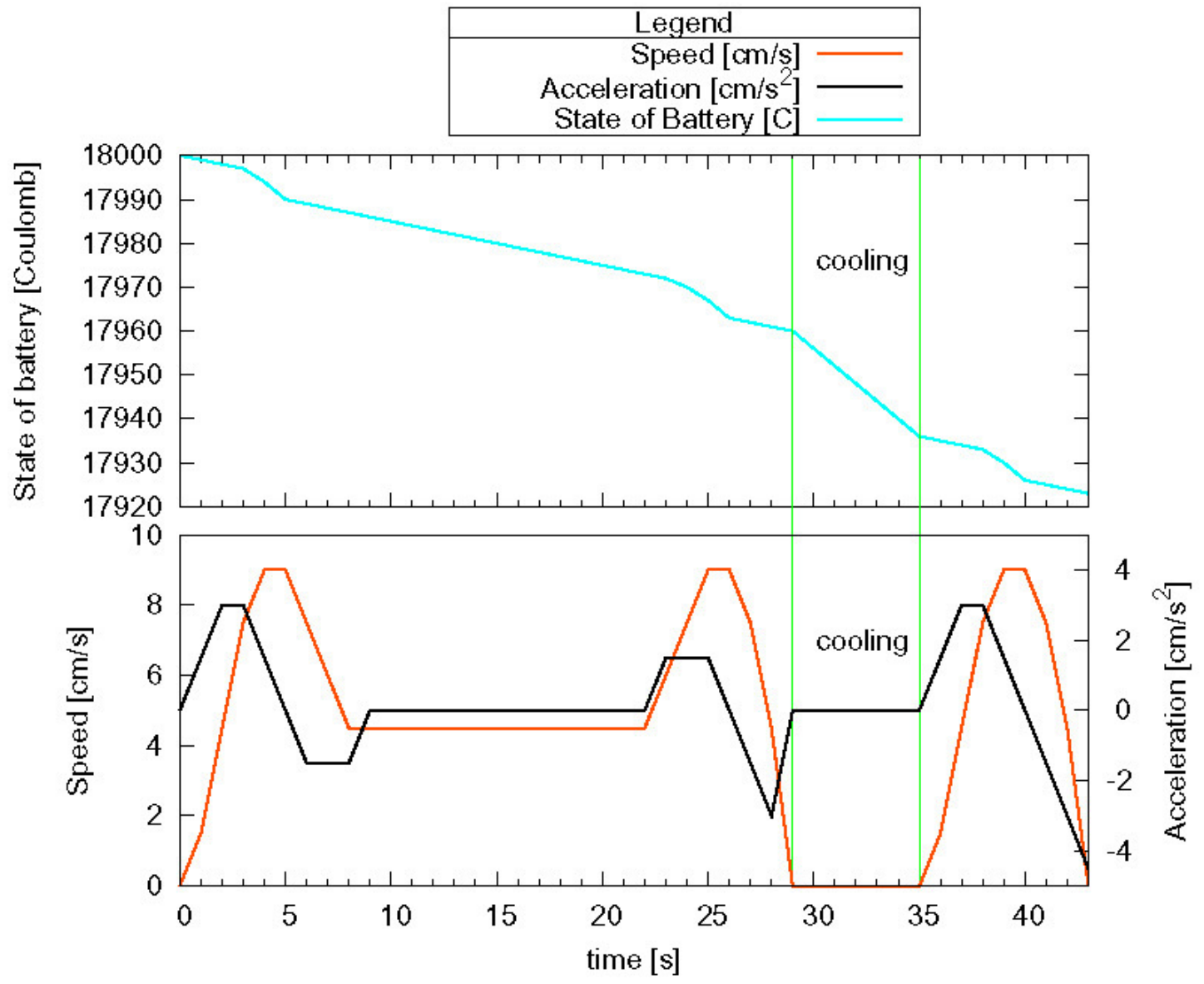

Figure 3. Optimal plan evolution: battery charge, speed and acceleration.

Finally, the generated plan has been further validated by simulating its execution on the rover model. The graphs in Figure 3 show the evolution of some important rover state variables during the simulation, which ends correctly after $d_{final} = 2\,m$. In particular, we can compare the battery discharge graph with the rover speed and acceleration during the entire course. It is worth noting that, in the highlighted cooling phase, the battery discharge rate is higher even if the vehicle is stopped, due to the instruments activation. Moreover, the peak speed of the rover is low enough to not exceed $v_{max}$ even in the presence of approximation errors up to 0.1, which is our discretisation threshold. Therefore, this is a completely safe plan.

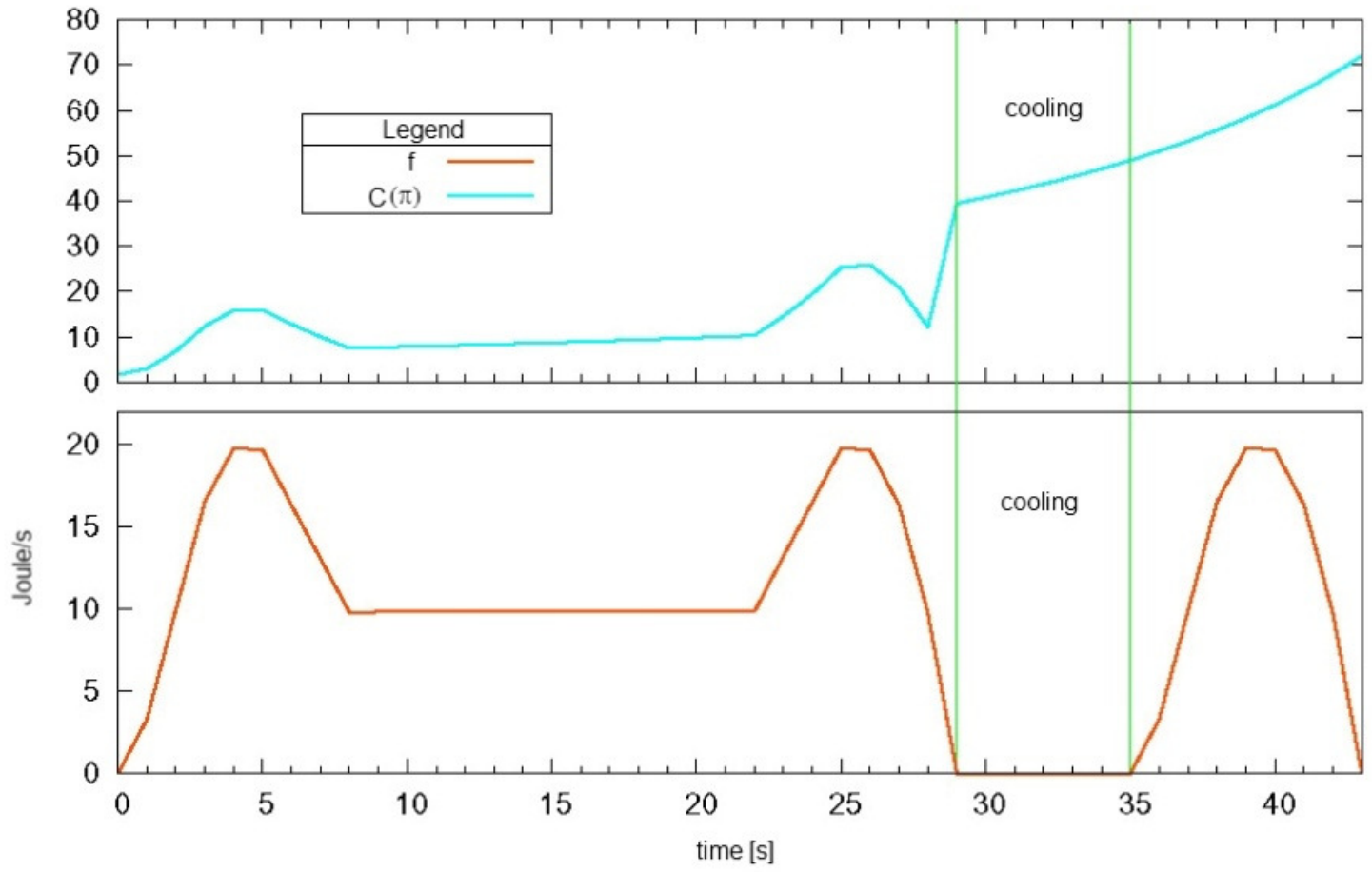

Figure 4. Optimal plan evolution: engine energy requirements and cost function.

Another interesting plan analysis is given in Figure 4, where we plot the rover engine energy requirements, i.e., the value of $f$ in Equation 1, and the value of the cost function $C(\pi)$ during the plan evolution. The graph clearly shows that, as required, the plan cost is very tightly related to the energy consumption, since the battery charge is a critical resource, whereas the time has a considerably lower impact (for example, look at the small increment of the cost when the required energy is constant, between $T = 8$ and $T = 22$).

## 5. CONCLUSIONS

In this paper we have shown how model checking based planning, and in particular the *UPMurphi universal planner*, can be profitably exploited to generate time and resource-optimal plans to control the engine of an autonomous planetary vehicle during an exploration mission.

The general characteristics of this kind of vehicle, together with the environmental conditions and the mission constraints, often lead to hybrid models with nonlinear dynamics, which are difficult to manipulate for most of the other planning tools. Therefore, planning in this context is often performed using only semi-automatic or even manual techniques on simplified models, whose results need to be checked for correctness and safety through simulation. Obviously, calculating a resource-optimal plan is even harder.

The UPMurphi optimal planner simplifies this process by automatically handling the system in its full complexity. Thus, the application of UPMurphi can be helpful to plan the activities of complex, realistic models of autonomous vehicles.

Indeed, we are working to further enhance the tool capabilities in order to handle more complex dynamics equations, in particular the ones involving differential equations, and to implement advanced state space pruning heuristics. On the other hand, we are studying how to use the UPMurphi in *simulation mode* to *automatically* validate the generated plans, thus providing a meaningful plan analysis to the user, which may include, e.g., some of the plots shown in this paper.